# An Extension to Hough Transform Based on Gradient Orientation

Tomislav Petković and Sven Lončarić
University of Zagreb
Faculty of Electrical and Computer Engineering
Unska 3, HR-10000 Zagreb, Croatia
Email: {tomislav.petkovic.jr, sven.loncaric}@fer.hr

*Abstract*—The Hough transform is one of the most common methods for line detection. In this paper we propose a novel extension of the regular Hough transform. The proposed extension combines the extension of the accumulator space and the local gradient orientation resulting in clutter reduction and yielding more prominent peaks, thus enabling better line identification. We demonstrate benefits in applications such as visual quality inspection and rectangle detection.

## I. Introduction

The Hough Transform (HT) is a commonly used method for line detection that is successfully applied in a large range of vision problems, starting from specific applications in industrial and robotic vision and extending to a general unconstrained problem of line detection in natural images. The success of the HT is primarily based on recasting a complex global line detection problem into a simple task of finding local peaks (concentrations of votes) in some parameter space.

The HT was proposed by Paul V. C. Hough [1] in 1962 and was introduced to computer vision community by Duda and Hart [2]. A comprehensive review of the HT was given by Illingworth and Kittler [3] in 1989. Later research introduced a randomized Hough transform (RHT) and its variants [4], [5], [6], [7] that eliminate the need for the quantized parameter space. Other improvements include better and more robust peak detection [8], [9] and extraction of line length [10], [11], [8]. Most of the above mentioned HT variants use only coordinates of extracted edge points and disregard other data that is often extracted from the input image during the edge or ridge detection.

In this paper we propose to improve upon a HT extension first suggested by O'Gorman and Clowes [12] that uses the gradient orientation to place a limit on the range of line orientation $\theta$ in the $(\theta, \rho)$ parameter space; if the detected gradient orientation is $\theta_0$ then votes are only accumulated for the predetermined range $\Delta\theta$ around the $\theta_0$, the interval $\langle\theta_0 - \Delta\theta, \theta_0 + \Delta\theta\rangle$. We extend this approach by combining the gradient orientation with the extension of accumulator space that makes straight line parametrization non-unique, but, combined with the range limit on orientation $\theta$, offers advantages of further clutter reduction and yields more prominent peaks.

The paper is organized as follows: In Section II a brief review of HT is given. In Section III a proposed accumulator array extension is introduced. In Section IV some results are presented and discussed. We conclude in Section V.

## II. The Hough Transform

Hesse normal form of a straight line is

$$\vec{r} \cdot \hat{n} - \rho = 0, \tag{1}$$

where $\vec{r} = \vec{i}x + \vec{j}y$ is the location vector of the point $(x, y)$, $\hat{n} = \vec{i}\cos\theta + \vec{j}\sin\theta$ is the unit normal vector of the straight line and $\rho \geq 0$ is the distance to the origin. For the HT Eq. (1) is usually rewritten as

$$\rho = x\cos(\theta) + y\sin(\theta), \tag{2}$$

which defines a sinusoid in $(\theta, \rho)$ parameter space that corresponds to a point $(x, y)$ in the input image. For the HT a sinusoid defined by Eq. (2) is drawn in the parameter space for every edge point $(x, y)$. Straight lines present in the input image are given by $(\theta, \rho)$ coordinates of the local peaks in the parameter space.

The parameters $\theta$ and $\rho$ are usually limited to either $[-\frac{\pi}{2}, \frac{\pi}{2}] \times \langle-\infty, +\infty\rangle$ or $[-\pi, \pi] \times [0, +\infty\rangle$ intervals, i.e. to ranges that produce unique mapping. For real world images $\pm\infty$ limit of parameter $\rho$ is replaced by $\rho_{max}$ defined by the finite size of the image.

Points $(x, y)$ are selected by an edge detector, most often by thresholding a gradient of the input image. Therefore, in addition to point coordinates, the direction and magnitude of the gradient are also known. Let $g_x$ and $g_y$ be components of the gradient in $x$ and $y$ directions. The gradient direction vector is perpendicular to the local edge so, as shown in [12], the orientation $\theta$ can be estimated as

$$\theta \approx \operatorname{atan}\frac{g_y}{g_x}, \tag{3}$$

and all sinusoids in $(\theta, \rho)$ space may be drawn only for a small interval of angles centered around the estimate of Eq. (3), thus reducing the clutter.

## III. Proposed Accumulator Extension

We propose to extend the accumulator so $\theta \in [-\pi, \pi]$ and $\rho \in [-\rho_{max}, \rho_{max}]$, where $\rho_{max}$ is determined by the size of the input image. This extension makes the straight line parametrization in $(\theta, \rho)$ space non-unique: every straight line in the image is represented by two different points in the proposed accumulator range. However, those two representations have exactly opposite normal directions. For example, consider the line $y = -\frac{3}{4}x + 3$ shown in Fig. 1; two unit normals are $\hat{n}$, pointing away from the origin, and $-\hat{n}$, pointing toward the origin. The normal $\hat{n}$ corresponds to the equation $\frac{12}{5} = \frac{3}{5}x + \frac{4}{5}y$, while the opposite normal $-\hat{n}$ corresponds to the equation $-\frac{12}{5} = -\frac{3}{5}x - \frac{4}{5}y$, which defines the same straight line. The HT of the point $(1.44, 1.92)$ closest to the origin marked with dot in Fig. 1 is shown in Fig. 2; two dots in Fig. 2 correspond to two possible directions of the straight line normal.

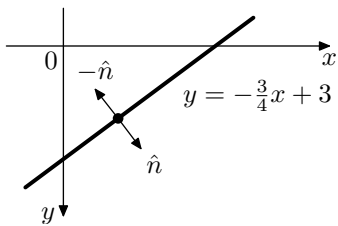

Fig. 1: A straight line $y = -\frac{3}{4}x + 3$ with two unit normals in opposite directions.

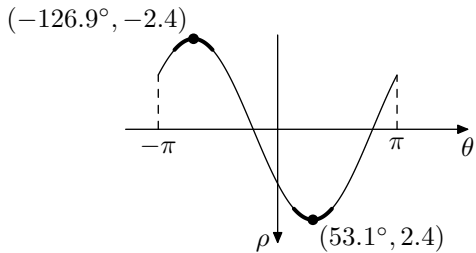

Fig. 2: HT of a point $(1.44, 1.92)$ on a line $y = -\frac{3}{4}x + 3$.

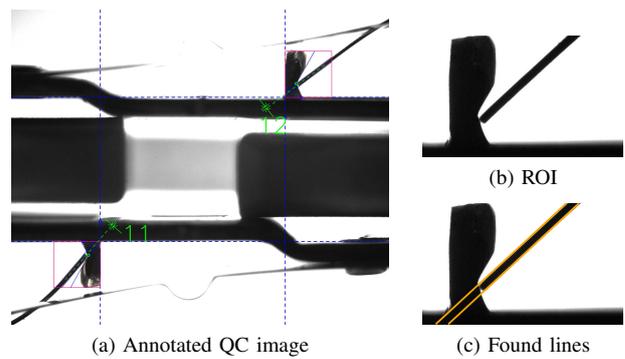

(a) Annotated QC image

(b) ROI

(c) Found lines

Fig. 3: QC example: Positioning of a flat spring in regard to a bearing structure must be examined. (a) is annotated QC image showing measures of interest. (b) shows a ROI. (c) shows found boundary lines delineating a flat spring.

However, the parametrization in proposed extended accumulator becomes *unique* if the notion of line direction is introduced; let the line direction vector $\hat{d}$ be defined so $\hat{n}$ and $\hat{d}$ form a right-hand coordinate system. Therefore, instead of straight lines we are detecting oriented straight lines.

The proposed extension depends on the gradient orientation. Let $g_x$ and $g_y$ be components of the gradient at $(x, y)$. The orientation $\theta$ for point at $(x, y)$ can be estimated as

$$\theta \approx \text{atan2}(g_y, g_x), \tag{4}$$

so the atan function of Eq. (3) yielding angles in the $[-\frac{\pi}{2}, \frac{\pi}{2}]$ interval is replaced by atan2 that yields angles in the $[-\pi, \pi]$ interval. The sinusoid in $(\theta, \rho)$ space is drawn only for a small interval of angles centered around the orientation estimate of Eq. (4), however, due to additional separation of points with opposing gradients we expect further clutter reduction in the HT domain.

Examining again the Fig. 2 demonstrates the difference: the plain HT would increment the accumulator array for all points on a sinusoid over $\theta \in [-\frac{\pi}{2}, \frac{\pi}{2}]$ range, the HT using the gradient direction to limit the angle range would increment the accumulator on the segment around $(53.1°, 2.4)$ point, and the proposed method would increment the accumulator on either segment around $(53.1°, 2.4)$ or around $(-126, 9°, -2.4)$ point, depending on the gradient orientation.

## IV. RESULTS AND DISCUSSION

In this Section we present several examples demonstrating the advantages of the proposed extension to the HT. For all input images the coordinate axes are as shown in Fig. 1 with the origin in the center. For all accumulator arrays the coordinate axes are as shown in Fig. 2 with the origin in the center. Values of accumulator arrays are mapped through a square root function, linearly scaled to available dynamic range and inverted; pure black corresponds to the highest accumulator value, white is zero, and mid-levels are gray. Such mapping compresses the dynamic range, reduces the intensities of peaks, and makes butterfly shapes around peaks clearly visible. Line distance to origin $\rho$ is measured in pixels and orientation $\theta$ in degrees. Range $\Delta\theta$ around $\theta_0$ was set to $22.5°$.

### A. Visual Quality Control

The HT is often applied in industrial vision tasks. We give two examples in visual quality control (QC) where a thin structure must be delineated.

The first example is QC of a thin flat spring whose position against the bearing structure must be inspected. An example is shown in Fig. 3: the input image is annotated showing structures of interest. The accumulator arrays for the regular HT and for the proposed HT are shown in Fig. 4: note the increased separation of two peaks that correspond to the upper and lower straight lines that delineate the thin flat spring. This separation effect is caused by the proposed accumulator extension and the use of the extracted gradient orientation; it will reduce clutter in the accumulator when spatially close features in the input image have different gradient orientations.

The second example demonstrates a more difficult visual QC task where position and inclination angles of a thin flat spring must be inspected. The flat spring is about one pixel thin, which is the main difficulty. The input image and steps of edge extraction are shown in Fig. 5: note the extreme thinness of the structure.

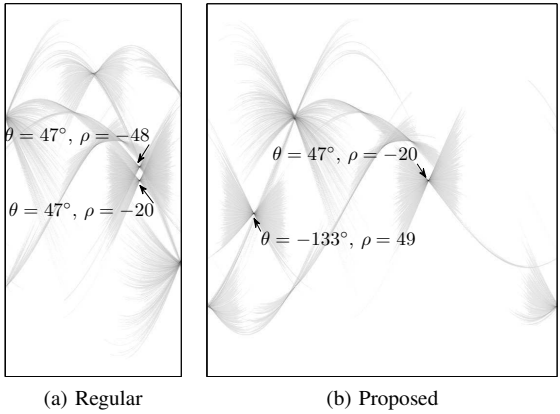

(a) Regular                    (b) Proposed

Fig. 4: Accumulator arrays for input image 3b. (a) is for regular HT where $-\frac{\pi}{2} < \theta < \frac{\pi}{2}$. (b) is for proposed HT where $-\pi < \theta < \pi$; note the improved separation of two marked lines.

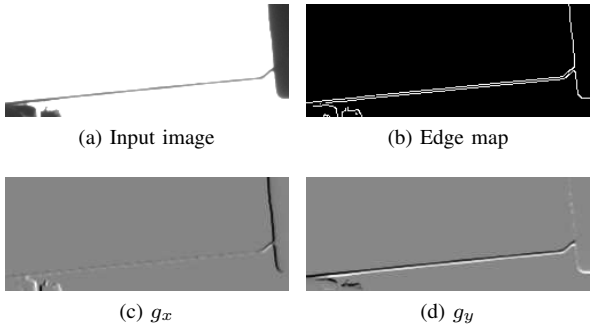

(a) Input image                    (b) Edge map

(c) $g_x$                    (d) $g_y$

Fig. 5: QC example: Examination of flat spring positioning. (a) is input image. (b) is edge map normally used HT. (c) and (d) are partial derivatives used to produce an edge map; two edges delineating the flat spring have clearly different $g_y$, but have the same properties in the edge map.

For regular HT an edge map shown in Fig. 5b is used. The problem with using only the edge map is that two edges corresponding to the upper and lower delineating straight line are close both in the image space and in the parameters space. Indeed, in the accumulator array shown in Fig. 6a two peaks are merged together and the separation of two delineating lines is not possible. Furthermore, the position of the peak corresponds neither to the upper nor to the lower delineating line, but is instead determined by the overlap of the butterfly shapes.

The proposed extension to the HT uses a gradient orientation in addition to the edge map. The gradient orientation is extracted from the gradient components $g_x$ and $g_y$ that are shown in Fig. 5, (c) and (d). Note that two edges of interest clearly have different properties when $g_x$ and $g_y$ are considered. Those differences are lost in the edge map where

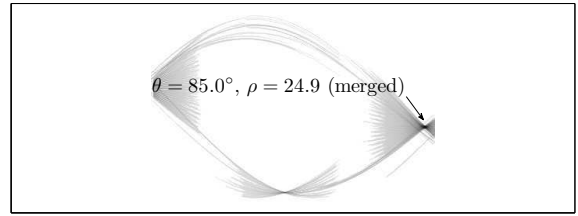

(a) Accumulator array for regular HT where $-\frac{\pi}{2} < \theta < \frac{\pi}{2}$

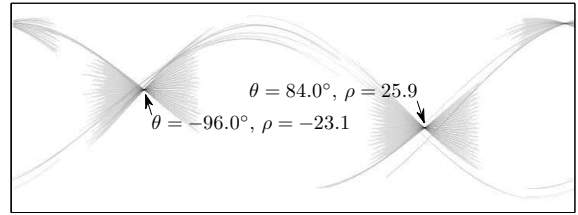

(b) Accumulator array for proposed HT where $-\pi < \theta < \pi$

Fig. 6: HT accumulator arrays for image 5a. Note merging of two peaks in (a) and a clear separation in (b).

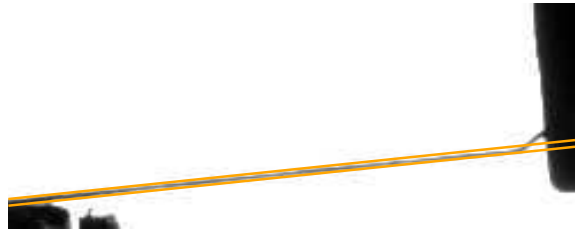

Fig. 7: Lines delineating the flat spring extracted using the proposed HT.

we observe only a combination $\sqrt{g_x^2 + g_y^2}$ (or $|g_x| + |g_y|$ for time-critical applications). Using the gradient orientation as proposed enables clear separation of two peaks as shown in Fig. 6 and extraction of both delineating lines is possible using the proposed accumulator array. The extracted lines using the proposed scheme are shown in Fig. 7.

For the two QC examples the gradient was computed using the Sobel operator, the edge map was computed using the Canny method of OpenCV [13] with 210 as the upper threshold, and the resolution of the accumulator was set to $\Delta\rho = 1$ and $\Delta\theta = 0.5°$.

### B. Rectangle Detection

Another interesting example is rectangle detection in the HT domain proposed by Jung and Schramm [14]. Their method applies the regular HT to a sliding window and examines constellations of four peaks in the accumulator array. When the window center is at rectangle center a four-peak two-pair constellation $\{\{(\theta_1, \rho_1), (\theta_2, \rho_2)\}, \{(\theta_3, \rho_3), (\theta_4, \rho_4)\}\}$ describing a rectangle satisfies the following properties:

1) peaks in a pair have the same orientation, $\theta_1 = \theta_2 = \alpha$ and $\theta_3 = \theta_4 = \beta$,

2) orientations $\alpha$ and $\beta$ are separated by 90° degrees,

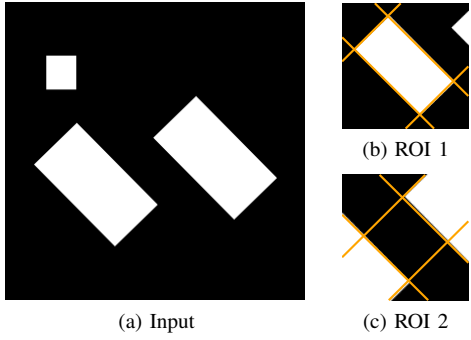

(b) ROI 1

(a) Input

(c) ROI 2

Fig. 8: A synthetic example for rectangle detection. Two ROIs are selected: ROI 1 (b) is centered around lower left rectangle and is an example of a proper response, and ROI 2 (c) is centered in the middle between two large rectangles and is an example of a false response.

3) two peaks of a pair have canceling distance to origin, $\rho_1 + \rho_2 = 0$ and $\rho_3 + \rho_4 = 0$,

4) two peaks of a pair have the same height (equal to lengths of rectangle's sides $a$ and $b$), and

5) the vertical distances within each pair are equal to lengths of rectangle's sides, $|\rho_1 - \rho_2| = a$ and $|\rho_3 - \rho_4| = b$.

In practice equality requirements are relaxed to absolute differences being less than chosen thresholds. The proposed HT allows expansion of the rectangle detection scheme by introducing the orientations of any of the four sides. The constellation properties change to:

1) peaks in a pair have the opposite orientations, $\theta_1 = \theta_2 + 180° = \alpha$ and $\theta_3 = \theta_4 + 180° = \beta \pmod{180°}$,

2) orientations $\alpha$ and $\beta$ are separated by $90°$ degrees,

3) signs of all four $\rho$'s are the same,

4) two peaks of a pair have equal distance to origin, $\rho_1 = \rho_2$ and $\rho_3 = \rho_4$,

5) two peaks of a pair have the same height (equal to lengths of rectangle's sides $a$ and $b$), and

6) absolute sum of distances $\rho$ within a pair is equal to lengths of rectangle's sides, $|\rho_1 + \rho_2| = a$ and $|\rho_3 + \rho_4| = b$.

A synthetic example image with two ROIs is shown in Fig. 8. A typical rectangle constellation for the regular HT is shown in Fig. 9a. For the proposed HT the constellation of Fig. 9a is transformed to a constellation in Fig. 9b: four peaks around the origin are unwrapped into a line, the distance between four peaks along the $\theta$ axis is $90°$, and all $\rho$'s have the same sign (gradients of all edges have directions either toward or away from the origin of the sliding window).

The original method [14] detects a false rectangle in ROI 2 (Fig. 8c) as the constellation shown in Fig. 10a matches. The proposed extension successfully eliminates such false rectangles as the constellation shown in Fig. 10b does not have all $\rho$'s with the same sign (gradients of two edges are toward and of other two are away from the origin of the sliding

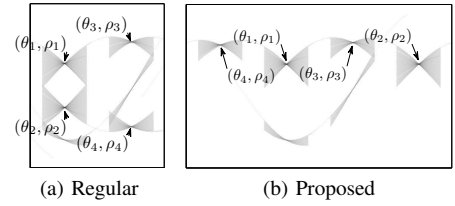

(a) Regular                    (b) Proposed

Fig. 9: Constellations describing a true rectangle corresponding to the sliding window shown in Fig. 8b.

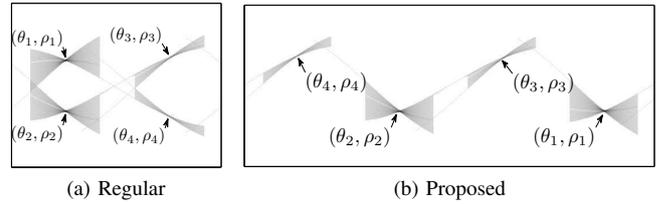

(a) Regular                    (b) Proposed

Fig. 10: Constellations describing a false rectangle corresponding to the sliding window shown in Fig. 8c. Note that the false rectangle will not be detected if proposed HT is used as the constellation (b) does not match.

window).

For two examples the gradient was computed using the Sobel operator, the edge map was computed using the Canny method of OpenCV [13] with 50 as the upper threshold, and the resolution of the accumulator was set to $\Delta\rho = 0.5$ and $\Delta\theta = 0.5°$.

### C. Discussion and Implementation

The proposed extension to the HT is simple and effective. It can be easily included into advanced Hough schemes such as RHT.

The characteristic butterfly shape around peaks in the parameter space remains the same so accurate and robust peak detection scheme [8] and line segment estimator [10] are directly applicable and may be used without further modifications.

Note that this approach is different than computing the regular Hough transform twice, once for bright-to-dark and once for dark-to-bright edges as it avoids the problem of the preferred direction, i.e. reversing the preferred direction transforms any bright-to-dark edge to a dark-to-bright edge thus making all edges orthogonal to the preferred direction difficult to detect.

We have implemented the proposed improved HT in C/C++. The implementation is based on OpenCV [13] and is freely available under BSD license at http://www.fer.unizg.hr/ipg/resources/HT.

### V. Conclusion

We proposed a simple and effective extension to the regular HT. The proposed extension combines the extension of the accumulator array and the local gradient orientation resulting in clutter reduction and yielding more prominent peaks, thus enabling better line identification. We have demonstrated benefits in applications such as visual quality inspection and rectangle detection.

## ACKNOWLEDGMENT

This research has been supported in part by the Research Centre for Advanced Cooperative Systems (EU FP7 #285939). The authors would like to thank Elektro-kontakt d.d. Zagreb, specifically mr. Ivan Tabaković, for provided images.

## REFERENCES

[1] P. V. C. Hough, "Method and means for recognizing complex patterns," U.S. Patent 3,069,654, Dec. 1962, United States Atomic Energy Commission.

[2] R. O. Duda and P. E. Hart, "Use of the Hough transformation to detect lines and curves in pictures," *Communications of the ACM*, vol. 15, no. 1, pp. 11–15, 1972.

[3] J. Illingworth and J. Kittler, "A survey of the Hough transform," *Computer vision, graphics, and image processing*, vol. 44, no. 1, pp. 87–116, 1988.

[4] L. Xu, E. Oja, and P. Kultanen, "A new curve detection method: randomized Hough transform (RHT)," *Pattern recognition letters*, vol. 11, no. 5, pp. 331–338, 1990.

[5] L. Xu and E. Oja, "Randomized Hough transform (RHT): basic mechanisms, algorithms, and computational complexities," *CVGIP Image Understanding*, vol. 57, pp. 131–131, 1993.

[6] J. Matas, C. Galambos, and J. Kittler, "Progressive probabilistic Hough transform." British Machine Vision Conference, 1998.

[7] L. Xu and E. Oja, "Randomized Hough transform," in *Encyclopedia of Artificial Intelligence*, J. R. R. Dopico, J. Dorado, and A. Pazos, Eds. IGI Global, 2008, pp. 1354–1361.

[8] Y. Furukawa and Y. Shinagawa, "Accurate and robust line segment extraction by analyzing distribution around peaks in Hough space," *Computer Vision and Image Understanding*, vol. 92, no. 1, pp. 1–25, 2003.

[9] S. Guo, T. Pridmore, Y. Kong, and X. Zhang, "An improved Hough transform voting scheme utilizing surround suppression," *Pattern Recognition Letters*, vol. 30, no. 13, pp. 1241–1252, 2009.

[10] M. Akhtar and M. Atiquzzaman, "Determination of line length using Hough transform," *Electronics Letters*, vol. 28, no. 1, pp. 94–96, 1992.

[11] M. Atiquzzaman and M. Akhtar, "A robust Hough transform technique for complete line segment description," *Real-Time Imaging*, vol. 1, no. 6, pp. 419–426, 1995.

[12] F. O'Gorman and M. Clowes, "Finding picture edges through collinearity of feature points," *IEEE Transactions on Computers*, vol. 25, no. 4, pp. 449–456, 1976.

[13] "OpenCV [Open Source Computer Vision] library," http://opencv.org/, Jan. 2014.

[14] C. R. Jung and R. Schramm, "Rectangle detection based on a windowed Hough transform," in *Computer Graphics and Image Processing, 2004. Proceedings. 17th Brazilian Symposium on*. IEEE, 2004, pp. 113–120.